\documentclass{article}

\usepackage{arxiv}

\usepackage[utf8]{inputenc} 
\usepackage[T1]{fontenc}    
\usepackage{hyperref}       
\usepackage{url}            
\usepackage{booktabs}       
\usepackage{amsfonts}       
\usepackage{nicefrac}       
\usepackage{microtype}      
\usepackage{lipsum}

\usepackage{bm}
\usepackage{amsmath}
\usepackage{amssymb}
\usepackage{amsfonts}
\usepackage{graphicx}
\def\vec#1{\mathbf{#1}}
\usepackage{algorithm}
\usepackage{algorithmic}
\usepackage{comment}

\title{Modeling Nonlinear Dynamics in Continuous Time with Inductive Biases on Decay Rates and/or Frequencies}

\author{
  Tomoharu Iwata\\
  NTT Communication Science Laboratories, Kyoto, Japan\\
  \AND
  Yoshinobu Kawahara\\
  Graduate School of Information Science and Technology, Osaka University, Osaka, Japan\\
  Center for Advanced Intelligence Project, RIKEN, Tokyo, Japan\\  
}
\date{}

\begin{document}
\maketitle

\begin{abstract}
  We propose a neural network-based model for nonlinear dynamics in continuous time that can impose inductive biases on decay rates and/or frequencies. Inductive biases are helpful for training neural networks especially when training data are small. The proposed model is based on the Koopman operator theory, where the decay rate and frequency information is used by restricting the eigenvalues of the Koopman operator that describe linear evolution in a Koopman space. We use neural networks to find an appropriate Koopman space, which are trained by minimizing multi-step forecasting and backcasting errors using irregularly sampled time-series data. Experiments on various time-series datasets demonstrate that the proposed method achieves higher forecasting performance given a single short training sequence than the existing methods.
\end{abstract}

\section{Introduction}

Analyzing and forecasting nonlinear dynamical systems are important
in a wide variety of fields, such as physics,
epidemiology, social science, and marketing.
Neural networks, such as recurrent neural networks~\cite{hochreiter1997long,che2018recurrent},
and neural ordinary differential equations (ODEs)~\cite{chen2018neural},
have been used for modeling black-box nonlinear dynamical systems given time-series data.
However, these models generally require many training data.
To alleviate such problems, inductive biases on the dynamical system
can be used for modeling.
For example, Hamiltonian neural networks
can model dynamics that obey exact conservation
laws~\cite{greydanus2019hamiltonian,course2021weak},
and monotonic neural networks can model monotonically increasing dynamics~\cite{archer1993application,sill1998monotonic,wehenkel2019unconstrained}.

In this paper, we propose a simple yet effective neural network-based method
for modeling black-box nonlinear dynamical systems in continuous time
that can impose inductive biases on (a part of) decay rates and/or frequencies.
The proposed model can be trained with a
small number of irregularly sampled time-series data.
With the decay rate, we can constrain the dynamics of models
to be conservative, damped, or diverging.
Many physical and biological systems are
known to be conservative or damped~\cite{janson2012non}.
Also, we can know frequencies in many dynamics.
For example, 
living things have a circadian rhythm of roughly every 24 hours,
human activities follow a weekly periodicity,
and climate data show an annual cycle.
Even when decay rates and frequencies are known,
its dynamics cannot be determined uniquely if the dynamics is nonlinear. 

The proposed model is based on
the Koopman operator theory~\cite{koopman1931hamiltonian,mezic2005spectral}.
With this theory,
a nonlinear dynamical system is lifted
to the corresponding linear one
in a possibly infinite-dimensional space,
which we call a Koopman space,
by embedding states using a nonlinear function.
We specify decay rates and frequencies of our models
by restricting the eigenvalues of a Koopman operator that
describe the evolution in the Koopman space
since the real parts of the eigenvalues characterize the decay rates,
and the imaginary parts characterize the frequencies.
For modeling with the Koopman operator,
we need to find a Koopman space that is appropriate for the given time-series data.
We use encoder and decoder neural networks to find the Koopman space.
Although many neural network-based methods with the Koopman operator theory
have been proposed~\cite{takeishi2017learning,lusch2018deep,yeung2019learning,lee2020model,iwata2020neural,azencot2020forecasting,li2019learning,han2021desko},
no existing methods put constraints on decay rates or frequencies.

Existing neural network-based models in continuous time, such as neural ODEs,
require high computational cost for training
since they need to backpropagate through an ODE solver
or solve an adjoint ODE for each training epoch.
On the other hand,
the proposed model can analytically obtain a solution of
the ODE and its derivative
in constant time with respect to the forecasting period
due to the linearity of evolution in the Koopman space,
which enables efficient training.
We can make a prediction at past time points (backcast)
using the same model for prediction at future time points (forecast)
without additional parameters
due to the reversibility of the linear Koopman operator.
Therefore, we can augment training data
by adding backcast errors to the training objective function
of forecast errors.
The data augmentation is beneficial especially when
training data are small.


\section{Related work}
\label{sec:related}

A number of neural networks have been proposed
that can use the knowledge on dynamical systems~\cite{karniadakis2021physics}.
Physics-informed neural networks~\cite{raissi2019physics,takeishi2021physics}
train models such that they satisfy given data
while respecting given differential equations.
Unlike the proposed method,
they need explicit forms of differential equations.
Hamiltonian and generalized Hamiltonian neural networks
can place physics-inspired priors on systems~\cite{greydanus2019hamiltonian,course2021weak}.
Although learning stable dynamics models has been studied~\cite{khansari2011learning,neumann2013neural,umlauft2017learning,duncker2019learning,chang2019neural,kolter2019learning,massaroli2020stable,takeishi2021learning},
these methods cannot place priors about frequencies.

Neural network-based methods for modeling continuous-time ODEs
require derivative regression~\cite{greydanus2019hamiltonian}
or computationally expensive
numerical integration to solve the ODEs~\cite{chen2018neural,rubanova2019latent,toth2019hamiltonian,matsubara2021symplectic,chen2021eventfn}.
Although derivative regression is efficient,
it needs the approximation of the derivatives
by finite difference~\cite{chartrand2011numerical},
which is susceptible to noise in data.
Weak form learning~\cite{schaeffer2017sparse,course2021weak}
has been proposed for efficient training of neural ODEs.
However, it requires that the time measurements
are sufficiently close together.
Neural networks can be used for forecasting values
in continuous time by additionally inputting the time
interval information~\cite{che2018recurrent}.
However, they cannot put constraints on decay rates and frequencies with the trained model.

Many models for periodic dynamics have been proposed,
which include autoregressive models~\cite{vecchia1985maximum,dudek2016periodic}
and neural networks~\cite{zhang2020periodic}.
However, they cannot impose priors on specific frequencies.
Some methods use periodic functions such as Fourier series
of given frequencies for modeling
periodicity~\cite{fidino2017using,smyl2020hybrid,taylor2018forecasting}.
However, they are for discrete time, and
cannot use the knowledge on decay rates.
\cite{lusch2018deep} proposed neural network-based models
that parameterize the eigenvalues of a Koopman operator with Jordan blocks.
However, they estimate the eigenvalues,
and do not impose the eigenvalues as inductive biases.
A method was proposed to learn a Koopman operator with a regularizer that softly constrains the eigenvalues~\cite{iwata2020neural}.
Since it is soft constraints,
the learned models do not necessarily satisfy the constraints.
On the other hand, the proposed method can learn models
that always satisfy the given constraints,
and it does not need hyperparameters for regularizers to be tuned.
\cite{ohnishi2021koopman} proposed a method to
control the eigenvalues of systems with exogenous input,
which is not for modeling the dynamics given observed time-series data.
Generalized Laplace average (GLA)~\cite{budivsic2012applied,mohr2014construction} is a model-based approach for finding Koopman eigenfunctions given eigenvalues and a dynamical system, i.e., differential equations to describe the
change in time.
Therefore, GLA cannot be used when observed time-series data are given instead of a dynamical system.
On the other hand, the proposed method is a data-driven approach that learns
a black-box dynamical system from time-series data.

\section{Preliminaries: Koopman operator theory}
\label{sec:preliminaries}

We briefly review the Koopman operator theory in this section.
We consider nonlinear continuous-time dynamical system,
$\frac{d\vec{x}(t)}{dt}=f(\vec{x}(t))$,
where $\vec{x}(t)\in\mathcal{X}$ is the state at time $t$.
Denote by $F^\tau$ the flow induced by the continuous-time system for time period $\tau$.
Then, the family of Koopman operators $\mathcal{K}^\tau$ associated with $F^\tau$ is defined as an infinite-dimensional linear operator that acts on observables
$g:\mathcal{X}\rightarrow\mathbb{R}$ (or $\mathbb{C}$)~\cite{koopman1931hamiltonian},
$\mathcal{K}^\tau g = g(F^\tau(\vec{x}(t)))$,
with which the analysis of nonlinear dynamics
can be lifted to a linear regime.
If the Koopman semigroup of operators is strongly continuous~\cite{engel2001one}, the limit
$\lim_{\tau\to 0}\frac{\mathcal{K}^\tau g-g}{\tau}\triangleq \mathcal{K} g$
exists, which defines the infinitesimal Koopman generator $\mathcal{K}$. Since the generator has the relation $\mathcal{K}g = f\cdot \nabla g$, where $\nabla$ denote the gradient operator, we have $\frac{d g(x(t))}{dt} = \mathcal{K} g(x(t))$.
When $\mathcal{K}$ has only discrete spectra,
observable $g$ is expanded by the eigenfunctions of $\mathcal{K}$,
$\varphi_{k}:\mathcal{X}\rightarrow\mathbb{C}$,
$g(\vec{x}(t))=\sum_{k=1}^{\infty}\alpha_{k}\varphi_{k}(\vec{x}(t))$,
where $\alpha_{k}\in\mathbb{C}$ is the coefficient.
Then the dynamics of observable $g$ is factorized,
\begin{align}
  g(\vec{x}(t))
  =\sum_{k=1}^{\infty}\exp(t\lambda_{k})\alpha_{k}\varphi_{k}(\vec{x}(0)),
\label{eq:koopman0}
\end{align}
where $\lambda_{k}\in\mathbb{C}$ is the eigenvalue of eigenfunction $\varphi_{k}$.
Since $\lambda_{k}$ is the only time dependent factor in the right-hand side of Eq.~(\ref{eq:koopman0}),
$\lambda_{k}$ characterizes the time evolution.
In particular,
the exponential of its real part $\exp(\mathrm{Re}(\lambda_{k}))$
determines the decay rate,
and its imaginary part $\mathrm{Im}(\lambda_{k})$
determines the frequency.

Although the existence of the Koopman operator is theoretically guaranteed in various situations, its practical use is limited by its infinite dimensionality.
We can assume the restriction of $\mathcal{K}$
to a finite-dimensional subspace $\mathcal{G}$.
If $\mathcal{G}$ is spanned by a finite number of functions $\{g_{1},\dots,g_{K}\}$,
then the restriction of $\mathcal{K}$ to $\mathcal{G}$,
which we denote $\vec{K}\in\mathbb{R}^{K\times K}$,
becomes a finite-dimensional operator,
$\frac{d\vec{g}(t)}{dt}=\vec{K}\vec{g}(t)$,
where $\vec{g}(t)=[g_{1}(\vec{x}(t)),\dots,g_{K}(\vec{x}(t))]\in\mathbb{R}^{K}$ is
a Koopman embedding vector at time $t$.

\section{Proposed method}
\label{sec:proposed}

\subsection{Problem formulation}

We are given a sequence of measurement vectors $\{(\vec{y}_{n},t_{n})\}_{n=1}^{N}$,
where $\vec{y}_{n}\in\mathbb{R}^{M}$ is the $n$th measurement vector at continuous time $t_{n}\in\mathbb{R}$,
$t_{n+1}>t_{n}$, and $N$ is the length of the sequence.
It can be an unevenly, or irregularly, observed sequence, i.e., $t_{n+1}-t_{n}\neq t_{n'+1}-t_{n'}$.
We are also given decay rates and/or frequencies of (a part of) the dynamics.
Let $\vec{r}^{*}=[r^{*}_{1},\dots,r^{*}_{K_{\mathrm{r}}}]$ be the logarithm of the given decay rates,
and $\bm{\omega}^{*}=[\omega^{*}_{1},\dots,\omega^{*}_{K_{\mathrm{w}}}]$ be the given frequencies,
where $r^{*}_{k}\in\mathbb{R}$ and $\omega^{*}\in[0,2\pi)$.
For example, when the dynamics is known to obey conservation laws, we set $\vec{r}^{*}=\vec{0}$.
When the dynamics has daily and weekly patterns, we set
$\bm{\omega}^{*}=2\pi[1,1/7]$
with the one-day unit time.
Our aim is to learn a model of the continuous-time dynamics,
which can forecast measurement vector $\vec{y}(t)$ at future time $t>t_{N}$.
The proposed method is straightforwardly extended when a set of sequences are given,
$\{\{(\vec{y}_{dn},t_{dn})\}_{n=1}^{N_{d}}\}_{d=1}^{D}$, from a dynamical system,
where $d$ is the index of a sequence,
and $D$ is the number of sequences.

\subsection{Model}

We consider the following continuous-time nonlinear dynamical system,
\begin{align}
  \frac{d\vec{x}(t)}{dt}=f(\vec{x}(t)), \quad \vec{y}(t)=h(\vec{x}(t)),
\end{align}
where $\vec{x}(t)$ is the state at time $t$
that evolves by a black-box nonlinear function $f$,
and measurement vector $\vec{y}(t)$ is generated from the state
by a black-box nonlinear function $h$.
We embed measurement vector $\vec{y}(t)$ into the Koopman space
using encoder $\psi$,
\begin{align}
  \vec{g}(t)=(g\circ h^{-1})(\vec{y}(t))=\psi(\vec{y}(t)),
  \label{eq:gt}
\end{align}
where $\vec{g}(t)\in\mathbb{R}^{K}$ is the Koopman embedding vector at time $t$,
$K$ is the dimension of the Koopman space,
and $\psi$ is an encoder modeled by a neural network.

In the Koopman space, a linear dynamics is assumed as described in Section~\ref{sec:preliminaries},
\begin{align}
  \frac{d\vec{g}(t)}{dt}=\vec{K}\vec{g}(t),
  \label{eq:koopman}
\end{align}
where $\vec{K}\in\mathbb{R}^{K\times K}$ is the Koopman matrix.
We parameterize Koopman matrix $\vec{K}$ with
the following eigen decomposed structure,
\begin{align}
  \vec{K}=\vec{V}\bm{\Lambda}\vec{V}^{-1},
  \label{eq:eigendecomposition}
\end{align}
where $\bm{\Lambda}=\mathrm{diag}(\lambda_{1},\dots,\lambda_{K})\in\mathbb{C}^{K\times K}$
is a diagonal matrix of the eigenvalues,
$\lambda_{k}$ is the $k$th eigenvalue,
$\vec{V}=[\vec{v}_{1},\dots,\vec{v}_{k}]\in\mathbb{C}^{K\times K}$ is a set of eigenvectors,
and $\vec{v}_{k}$ is the $k$th eigenvector.
When Koopman matrix $\vec{K}$ has $K$ linearly independent eigenvectors,
we can decompose it as in Eq.~(\ref{eq:eigendecomposition}).
When we model a dynamics with an undiagonalizable Koopman matrix,
we can use Jordan canonical forms.

The real part of the eigenvalue represents the decay rate,
and the imaginary part represents the frequency.
The complex-valued eigenvalues always occur in complex conjugate pairs.
We parameterize the eigenvalues with real-valued parameters
$\vec{r}=[r_{1},\dots,r_{\lceil K/2\rceil}]\in\mathbb{R}^{\lceil K/2\rceil}$ and
$\bm{\omega}=[\omega_{1},\dots,\omega_{\lfloor K/2 \rfloor}]\in\mathbb{R}^{\lfloor K/2\rfloor}$
as follows,
\begin{align}
  \lambda_{2k-1}=r_{k}+i\omega_{k},\quad\lambda_{2k}=r_{k}-i\omega_{k},
  \label{eq:eigenvalue}
\end{align}
for $k=1,\dots,\lfloor K/2\rfloor$.
When $K$ is an odd number,
since $\vec{K}$ has at least one real-valued eigenvalue,
we parameterize the last eigenvalue by $\lambda_{K}=r_{\lceil K/2\rceil}$.
We fix (the part of) $\vec{r}$ and/or $\bm{\omega}$
with given $\vec{r}^{*}$ and/or $\bm{\omega}^{*}$,
$r_{1}=r_{1}^{*},\dots,r_{K_{\mathrm{r}}}=r_{K_{\mathrm{r}}}^{*}$ and
$\omega_{1}=\omega_{1}^{*},\dots,\omega_{K_{\mathrm{w}}}=\omega_{K_{\mathrm{w}}}^{*}$ while training.
The remaining ones, $r_{K_{\mathrm{r}}+1},\dots,r_{K}$ and $\omega_{K_{\mathrm{w}}+1},\dots,\omega_{K}$,
are parameters to be trained.
Even when a specific decay rate is unknown,
when we know that the dynamics is decaying,
we can parameterize the real parts of eigenvalues by $r_{k}=-\exp(r'_{k})$ with trainable parameter $r'_{k}$
such that they always give negative values.
Similarly, when we know that the dynamics is diverging,
we can parameterize them by $r_{k}=\exp(r'_{k})$.
When we know decay rates and/or frequencies are in a specific range between
$r_{\mathrm{start}}$ and $r_{\mathrm{end}}$,
we can parameterize them by
\begin{align}
  r_{k}=r_{\mathrm{start}}+\frac{r_{\mathrm{end}}-r_{\mathrm{start}}}{1+\exp(-r'_{k})},
  \label{eq:range}
\end{align}
using the sigmoid function such that they always give values within the range.
Since our model decomposes dynamics into multiple components with different decay rates and frequencies,
it works even when some of the components of the dynamics are different from the specified decay rates and frequencies.

The eigenvectors corresponding to a complex conjugate pair of eigenvalues
are also complex conjugate.
We parameterize the eigenvectors using real-valued parameters
$\vec{U}=[\vec{u}_{1},\dots,\vec{u}_{\lceil K/2 \rceil}]\in\mathbb{R}^{\lceil K/2 \rceil}$ and
$\vec{Z}=[\vec{z}_{1},\dots,\vec{z}_{\lfloor K/2 \rfloor}]\in\mathbb{R}^{\lfloor K/2 \rfloor}$
as follows,
\begin{align}
  \vec{v}_{2k-1}=\vec{u}_{k}+i\vec{z}_{k},\quad \vec{v}_{2k}=\vec{u}_{k}-i\vec{z}_{k},
  \label{eq:v}
\end{align}
for $k=1,\dots,\lfloor K/2\rfloor$, and
$\vec{v}_{K}=\vec{u}_{\lceil K/2\rceil}$ when $k$ is an odd number.
By the parameterizations in Eqs.~(\ref{eq:eigenvalue},\ref{eq:v}),
the constraints on conjugacy of eigenvalues and eigenvectors are
always satisfied while the number of parameters to be estimated is halved.

Given Koopman embedding $\vec{g}(t)$,
the Koopman embedding after time period $\tau\in\mathbb{R}$
is analytically calculated by
\begin{align}
  \hat{\vec{g}}(t+\tau|\vec{g}(t))=\vec{V}\exp(\tau\bm{\Lambda})\vec{V}^{-1}\vec{g}(t),
  \label{eq:gttau}
\end{align}
by solving ordinary differential equation
$\frac{d\vec{g}(t)}{dt}=\vec{V}\bm{\Lambda}\vec{V}^{-1}\vec{g}(t)$
using Eqs.~(\ref{eq:koopman},\ref{eq:eigendecomposition})
due to the linearity of the dynamics in the Koopman space.
Since the Koopman matrix is parameterized with the eigen decomposed structure as
in Eq.~(\ref{eq:eigendecomposition}),
Koopman embedding forecasting in Eq.~(\ref{eq:gttau}) is efficiently performed
only by multiplying period $\tau$ to eigenvalues $\bm{\Lambda}$
in constant time with respect to period $\tau$.

\begin{figure}[t!]
  \centering
  \includegraphics[width=23em]{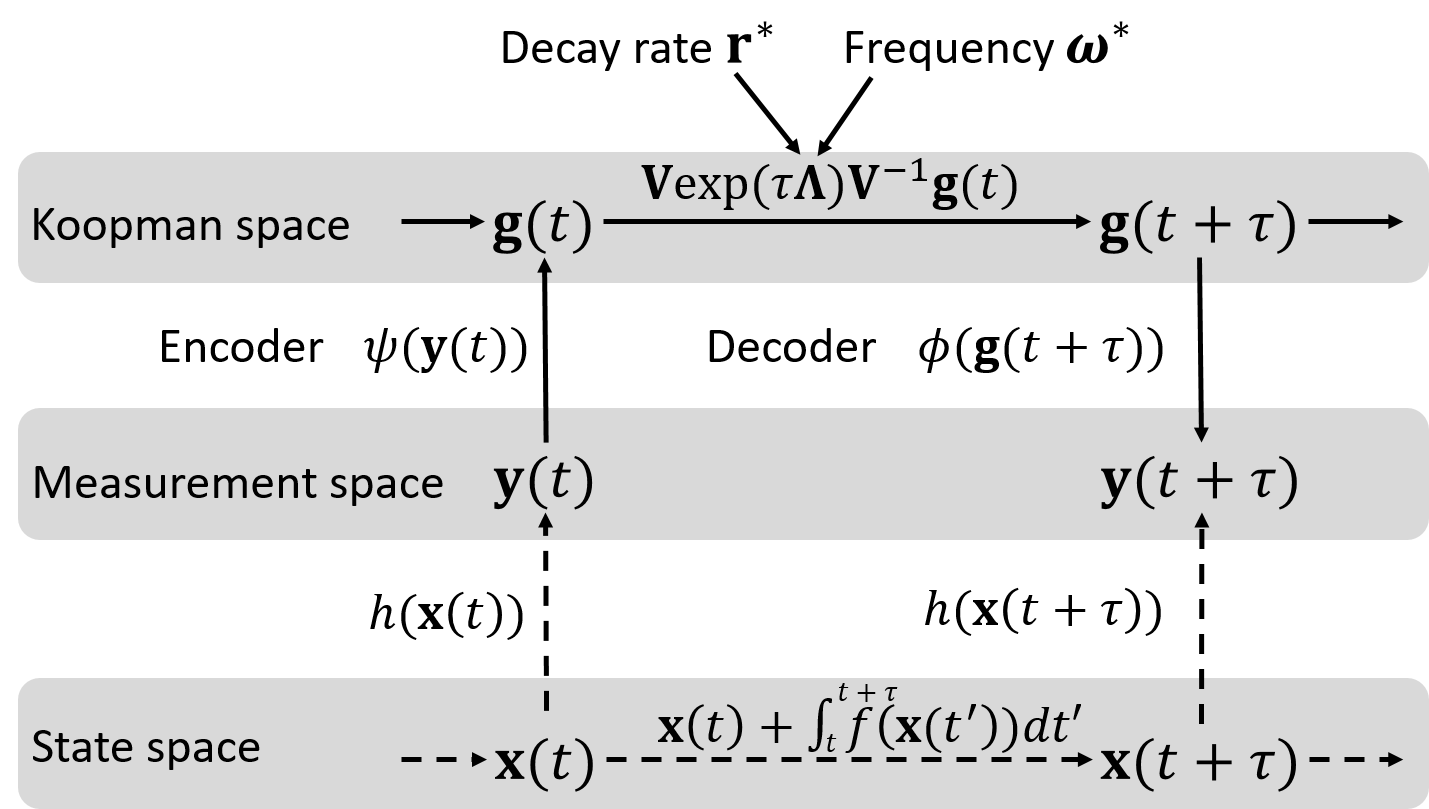}
  \caption{Our model (solid arrows) that approximates a black-box dynamical system (dashed arrows). With the black-box system, state $\vec{x}(t)$ is evolved by $\vec{x}(t)+\int_{t'=t}^{\tau}f(\vec{x}(t'))dt'$, and measurement vector $\vec{y}(t)$ is generated by $h(\vec{x}(t))$. Our model embeds measurement vector $\vec{y}(t)$ to the Koopman space by encoder $\psi$. Koopman embedding $\vec{g}(t)$ is evolved by $\vec{V}\exp(\tau\bm{\Lambda})\vec{V}^{-1}\vec{g}(t)$, where eigenvalues $\bm{\Lambda}$ are restricted by given decay rates $\vec{r}^{*}$ and frequency $\bm{\omega}^{*}$. Measurement vector $\vec{y}(t+\tau)$ is obtained from Koopman embedding $\vec{g}(t+\tau)$ using decoder $\phi$.}
  \label{fig:model}
\end{figure}

For obtaining measurement vectors from Koopman embeddings,
we use neural network-based decoder $\phi$,
\begin{align}
  \hat{\vec{y}}(t)=\phi(\vec{g}(t)).
  \label{eq:decoder}
\end{align}
Using Eqs.~(\ref{eq:gt},\ref{eq:gttau},\ref{eq:decoder}),
the predicted measurement vector after time period $\tau$ given $\vec{y}(t)$
is obtained by
\begin{align}
  \hat{\vec{y}}(t+\tau|\vec{y}(t))=\phi\left(
  \vec{V}\exp(\tau\bm{\Lambda})\vec{V}^{-1}\psi(\vec{y}(t))
  \right).
  \label{eq:ytau}
\end{align}
Figure~\ref{fig:model} illustrates our model.

\subsection{Training}

The parameters to be trained are 
real parts of eigenvalues $\vec{r}=[r_{K_{\mathrm{r}}},\dots,r_{\lceil K/2\rceil}]$,
imaginary parts of eigenvalues $\bm{\omega}=[\omega_{K_{\mathrm{w}}},\dots,\omega_{\lfloor K/2 \rfloor}]$
except for given $\vec{r}^{*}$ and $\bm{\omega}^{*}$,
real parts of eigenvectors $\vec{U}=[\vec{u}_{1},\dots,\vec{u}_{\lceil K/2 \rceil}]$,
imaginary parts of eigenvectors $\vec{Z}=[\vec{z}_{1},\dots,\vec{z}_{\lfloor K/2 \rfloor}]$,
and parameters $\bm{\Theta}$ of encoder $\psi$ and decoder $\phi$.
We train them by minimizing the following prediction error,
\begin{align}
  E(\vec{r},\bm{\omega},\vec{U},\vec{Z},\bm{\Theta})
  =\sum_{\nu=\nu_{\mathrm{start}}}^{\nu_{\mathrm{end}}}\sum_{n=\max(n+\nu,1)}^{\min(N-\nu,N)}\parallel\hat{\vec{y}}(t_{n+\nu}-t_{n}|\vec{y}_{n})-\vec{y}_{n+\nu}\parallel^{2},
    \label{eq:E}
\end{align}
where $\nu$ represents the number of prediction steps.
When $\nu=0$, the prediction corresponds to the auto-reconstruction,
where the measurement vector at the same time point
is reconstructed using the encoder and decoder
through the Koopman space.
When $\nu>0$, it corresponds to a forecast, which predicts the measurement vector at a future time.
When $\nu<0$, it corresponds to a backcast, which predicts the measurement vector at a past time.
Since the proposed model can forecast and backcast
in a single model in Eq.~(\ref{eq:ytau}),
we can use a negative value for the start number of prediction steps
$\nu_{\mathrm{start}}$.
The backcast errors in the objective function
can implicitly augment training data, which improves
the performance especially when the training sequence is short.

\subsection{Discrete-time model}

We can also model a discrete-time dynamical system in a similar way.
With a discrete-time system,
the radius of the eigenvalue corresponds to the decay rate,
and the argument of the eigenvalue corresponds to the frequency.
Therefore, we parameterize the eigenvalues by
\begin{align}
  \lambda_{2k-1}=r_{k}(\cos(\omega_{k})+i\sin(\omega_{k})),\quad
  \lambda_{2k}=r_{k}(\cos(\omega_{k})-i\sin(\omega_{k})),
  \label{eq:eigenvalue_discrete}
\end{align}
instead of Eq.~(\ref{eq:eigenvalue}) in the continuous-time case.
The predicted measurement vector after $\tau$ timesteps given $\vec{y}(t)$
is obtained by
\begin{align}
  \hat{\vec{y}}(t+\tau|\vec{y}(t))=\phi\left(
  \vec{V}\bm{\Lambda}^{\tau}\vec{V}^{-1}\psi(\vec{y}(t))
  \right),
\end{align}
where $\vec{y}(t)$ is the measurement vector at timestep $t$.
Since $\bm{\Lambda}$ is a diagonal matrix,
the power of $\tau$ can be calculated efficiently by
an element-wise exponentiation, $\bm{\Lambda}^{\tau}=\mathrm{diag}(\lambda_{1}^{\tau},\dots,\lambda_{K}^{\tau})$.

\section{Experiments}
\label{sec:experiments}

\begin{table*}[t]
  \centering
  \caption{Averaged test mean squared forecast errors and their standard errors.}
  \label{tab:error}
 {\tabcolsep=0.5em\begin{tabular}{lrrrrrr}
 \hline
 &Pendulum & PendulumI & VanDerPol &Fluid & SST & Bike\\
 \hline
 Ours & {\bf 0.713$\pm$0.355} & {\bf 0.476$\pm$0.216} & 1.207$\pm$0.171 & {\bf 0.187$\pm$0.036} & 0.784$\pm$0.076 & 0.854$\pm$0.074 \\
 OursF & - & - & {\bf 0.239$\pm$0.022} & - & {\bf 0.317$\pm$0.068} & {\bf 0.470$\pm$0.034} \\
 NDMD & 2.210$\pm$0.857 & 0.783$\pm$0.336 & 2.516$\pm$0.284 & 0.577$\pm$0.214 & 1.501$\pm$0.224 & 0.854$\pm$0.074 \\
 CKA & 2.021$\pm$0.634 & 13.953$\pm$0.900 & 1.836$\pm$0.246 & 0.343$\pm$0.067 & 1.350$\pm$0.191 & 0.593$\pm$0.046 \\
 DMD & 2.983$\pm$0.412 & 4.807$\pm$0.931 & 15.971$\pm$12.135 & 0.457$\pm$0.109 & $>10^{15}$ & 1.255$\pm$0.101 \\
DMDF
& - & - &
0.302$\pm$0.025
& - &
0.393$\pm$0.060
&
1.135$\pm$0.096\\
FFNN & 1.879$\pm$0.638 & 9.375$\pm$0.894 & 1.448$\pm$0.233 & 0.213$\pm$0.044 & 1.237$\pm$0.141 & 0.726$\pm$0.070 \\
FFNNF 
& - & - &
1.256$\pm$0.208	
& - &
1.098$\pm$0.157	
&
0.674$\pm$0.067	\\
LSTM & 3.107$\pm$0.588 & 13.177$\pm$0.871 & 1.643$\pm$0.265 & 0.434$\pm$0.060 & 0.665$\pm$0.127 & 0.887$\pm$0.106 \\
LSTMF
& - & - &
1.085$\pm$0.196
& - & 
0.557$\pm$0.119
&
0.798$\pm$0.083\\
LEM & 5.437$\pm$1.156&
    12.861$\pm$0.749&
    1.233$\pm$0.190&
    0.367$\pm$0.044&
    1.181$\pm$0.150&
    0.797$\pm$0.057\\
 NODE & 59.523$\pm$28.442 & 11.672$\pm$1.704 & 6.035$\pm$1.831 & 1.305$\pm$0.277& 1.008$\pm$0.089& $>10^{10}$\\
 HNN & 48.835$\pm$18.166 & 198.795$\pm$86.726 & 149.947$\pm$104.829
 & 2.970$\pm$1.092 & 21.626$\pm$6.400 & 34.507$\pm$5.524\\
 GHNN & 20.304$\pm$8.650& 132.721$\pm$64.891& 45.164$\pm$40.397& 23.374$\pm$4.582 & 26.328$\pm$8.547 & 41.862$\pm$6.390\\
 \hline
        \end{tabular}}
\end{table*}

\begin{figure*}[t]
  \centering
  {\tabcolsep=-0em\begin{tabular}{cccc}
    \includegraphics[width=11.5em]{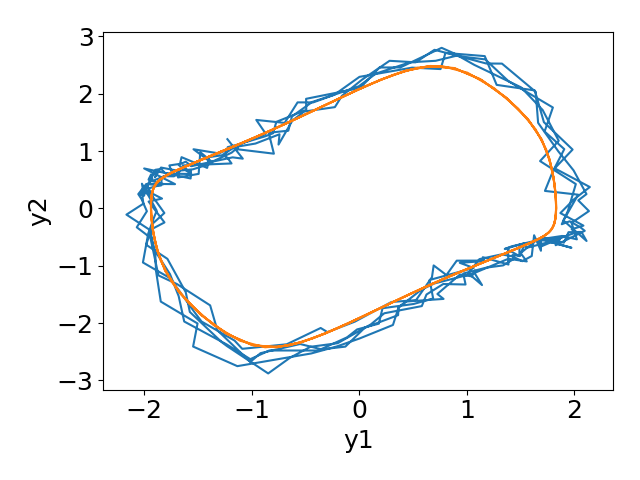}&
    \includegraphics[width=11.5em]{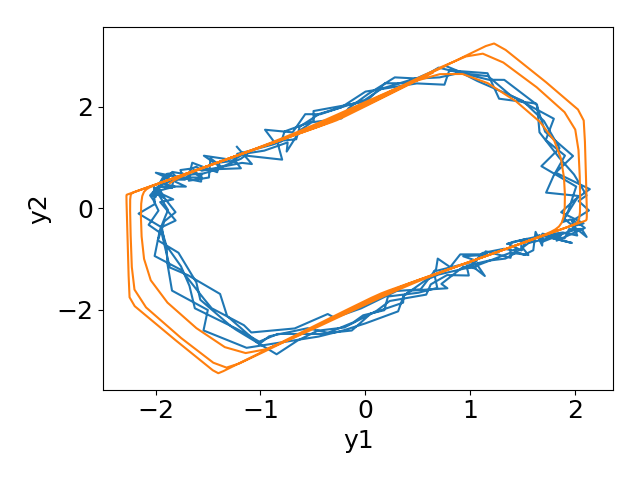}&
    \includegraphics[width=11.5em]{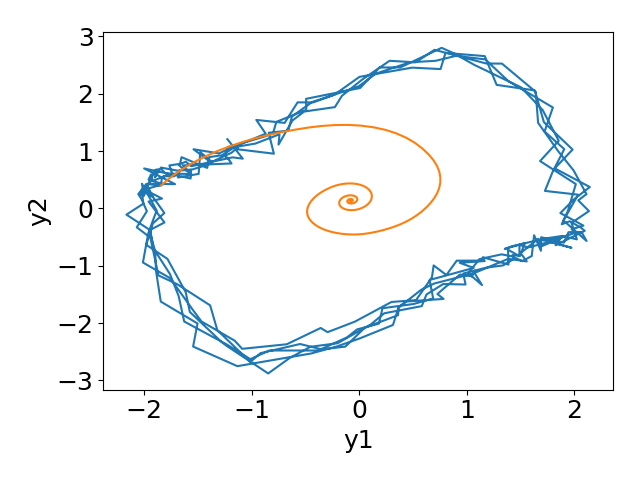}&
    \includegraphics[width=11.5em]{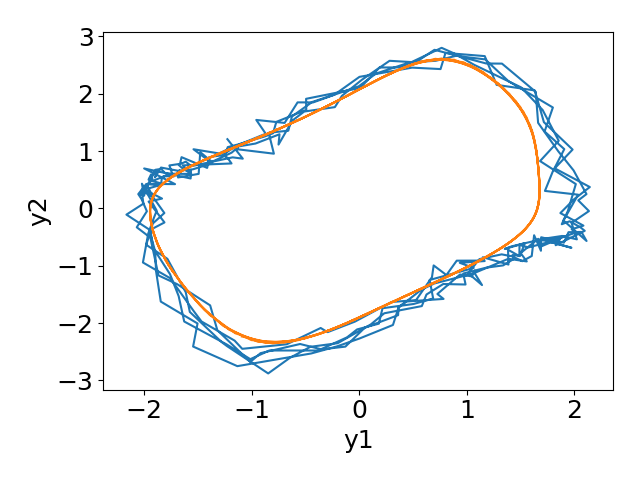}\\
    \includegraphics[width=11.5em]{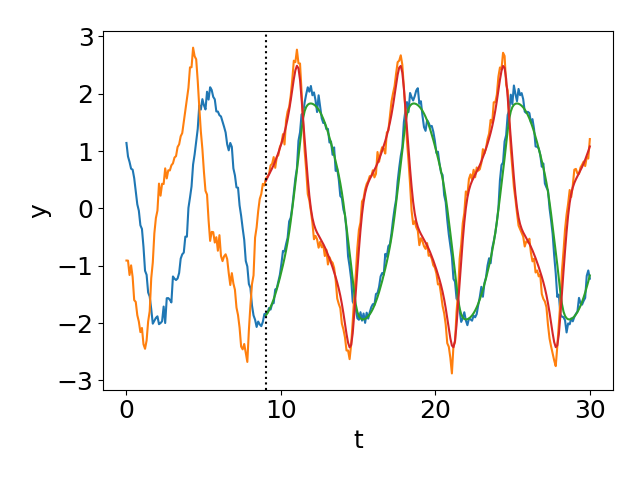}&
    \includegraphics[width=11.5em]{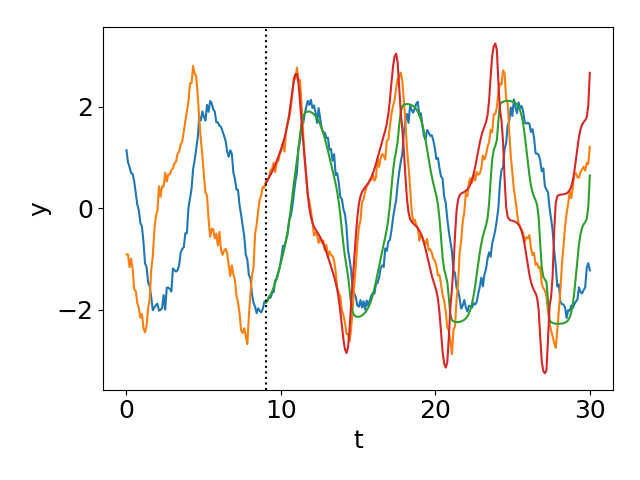}&
    \includegraphics[width=11.5em]{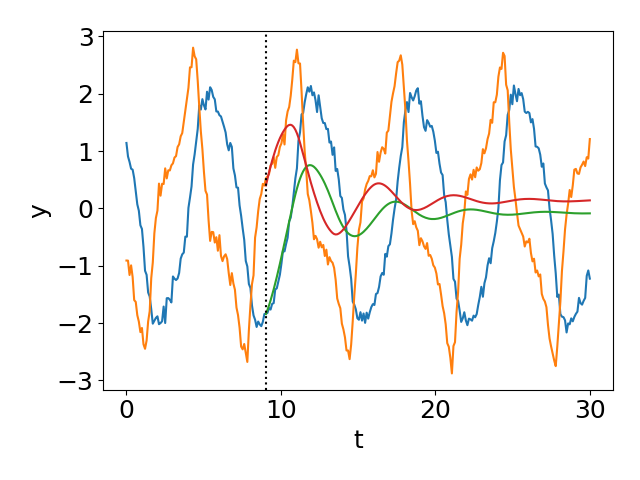}&
    \includegraphics[width=11.5em]{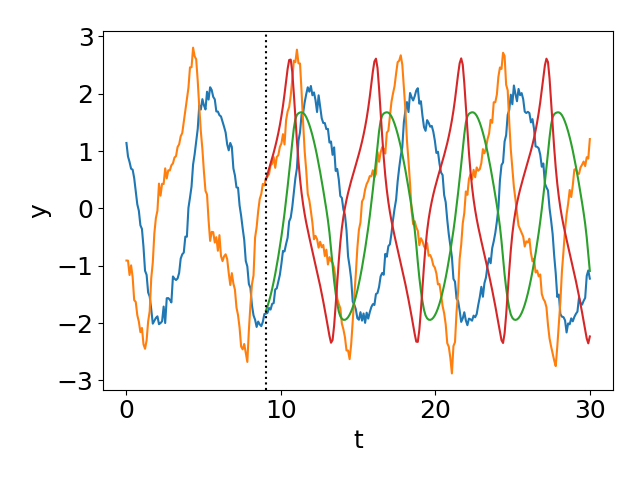}\\
    (a) OursF & (b) NDMD & (c) NODE & (d) HNN \\
\end{tabular}}  
  \caption{Predicted phase space (top rows) and time-series (bottom rows)
    on VanDerPol data. In the phase space, the horizontal axis is $y_{1}$,
    and the vertical axis is $y_{2}=\frac{dy_{1}}{dt}$, where the blue lines show the observed values, and the orange lines show the predicted values. In the time-series, the horizontal axis is the time, and the vertical axis is measurements $y_{1}$ and $y_{2}$, where the blue and orange lines show the observed values, and the red and green lines show the predicted values. The prediction starts from 10, which is shown by the vertical line.}
  \label{fig:vanderpol}
\end{figure*}

\begin{figure}[t]
  \centering
  \includegraphics[width=12.5em]{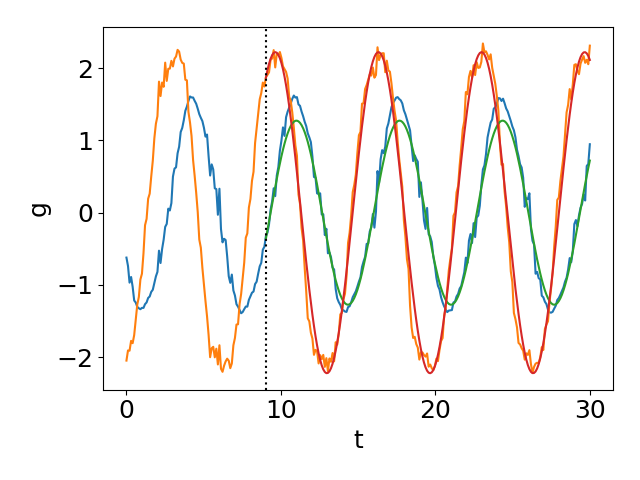}
  \caption{Time-series in the Koopman space on VanDerPol data.
    The horizontal axis is the time, and the vertical axis is the elements of the Koopman embedding vector $g_{1}$ and $g_{2}$, where the blue and orange lines show those of observed values, and the red and green lines show those of the predicted values.}
  \label{fig:vanderpol_g}  
\end{figure}

\begin{table*}[t]
  \centering
  \caption{Ablation study of the proposed method without backcast errors in the training objective function, and without structured eigenvectors. The values show the averaged test mean squared forecast errors and their standard errors on VanDerPol data.}
  \label{tab:ablation}
        {\tabcolsep=0.5em
          \begin{tabular}{lrrrrrr}
 \hline    
 & Pendulum &    PendulumI &    VanDerPol &    Fluid &    SST &    Bike \\
 \hline 
Ours & {\bf 0.713$\pm$0.355} & {\bf 0.476$\pm$0.216} & {\bf 1.207$\pm$0.171} & {\bf 0.187$\pm$0.036} &  {\bf 0.784$\pm$0.076} & {\bf 0.854$\pm$0.074} \\
w/o backcast & 1.522$\pm$0.675 & 1.485$\pm$0.667 & 1.475$\pm$0.227 & 0.437$\pm$0.126 & 1.168$\pm$0.090 & 1.440$\pm$0.132 \\
w/o struct eigen & 1.213$\pm$0.583 & 0.992$\pm$0.455 & 2.203$\pm$0.242 & 0.481$\pm$0.132 & 1.258$\pm$0.113 & 0.878$\pm$0.070 \\
 \hline
        \end{tabular}}
\end{table*}

\begin{table*}[t]
  \centering
  \caption{Averaged test mean squared forecast errors with different ranges of frequency information by the proposed method on VanDerPol data. When the range width is $b$ and the true frequency is $\omega^{*}$, the range is between $\omega_{\mathrm{start}}=\omega^{*}-b/2$ and $\omega_{\mathrm{end}}=\omega^{*}+b/2$.}
  \label{tab:range}
        {\tabcolsep=0.5em
          \begin{tabular}{lrrrrrr}
    \hline
    Range width & 0 & 0.001 & 0.003 & 0.01 & 0.03 & $\infty$ \\
    \hline
    Error & 0.239$\pm$0.022 & 0.257$\pm$0.023 & 0.259$\pm$0.023 & 0.306$\pm$0.033 & 0.601$\pm$0.098 & 1.207$\pm$0.171\\    
    \hline
  \end{tabular}}
\end{table*}

\begin{table}[t]
  \centering
  \caption{Averaged mean absolute errors of estimated frequency on VanDerPol data.}
  \label{tab:freq}
  {\tabcolsep=0.3em
  \begin{tabular}{rrrr}
    \hline
    Ours & NDMD & CKA & DMD \\
    \hline
{\bf 0.009$\pm$0.001} & 0.013$\pm$0.001 & 0.100$\pm$0.007 & 0.021$\pm$0.002\\
\hline
  \end{tabular}}
\end{table}

\begin{table*}[t]
  \centering
  \caption{Averaged mean absolute errors of estimated frequency on data with two frequencies.}
  \label{tab:freq2}
  {\tabcolsep=0.8em
  \begin{tabular}{rrrrr}
    \hline
    Ours & OursF & NDMD & CKA & DMD \\
    \hline
0.022$\pm$0.002 & {\bf 0.015$\pm$0.002} & 0.033$\pm$0.006 & 0.104$\pm$0.009 & 0.018$\pm$0.002\\
    \hline
    \end{tabular}}
\end{table*}

\begin{figure*}[t]
  \centering
  {\tabcolsep=0.3em\begin{tabular}{cccc}
    \multicolumn{4}{c}{(a) DMD with large data}\\
    \includegraphics[width=11em]{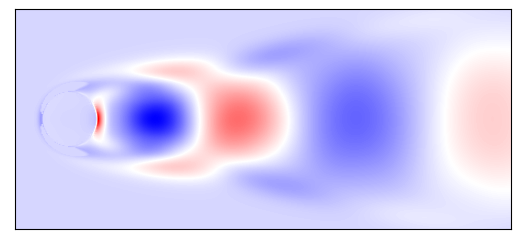}&    
    \includegraphics[width=11em]{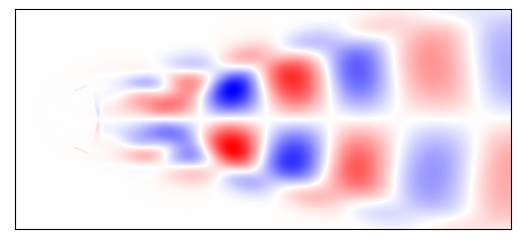}&
    \includegraphics[width=11em]{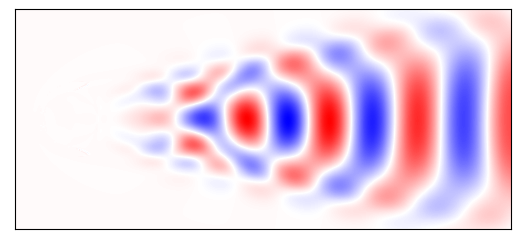}&
    \includegraphics[width=11em]{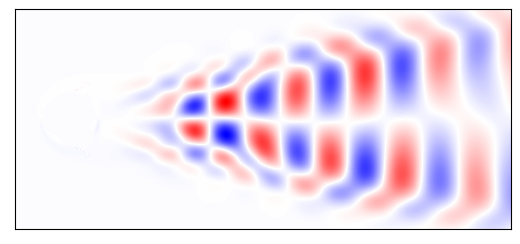}\\
  \multicolumn{4}{c}{(b) DMD with small data}\\
  \includegraphics[width=11em]{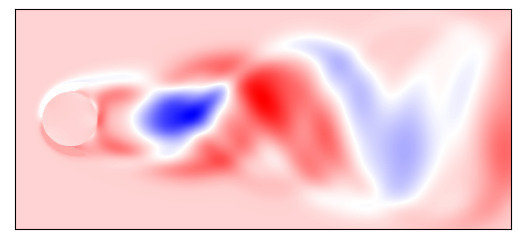}&
  \includegraphics[width=11em]{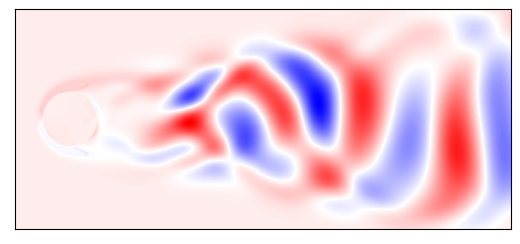}&
  \includegraphics[width=11em]{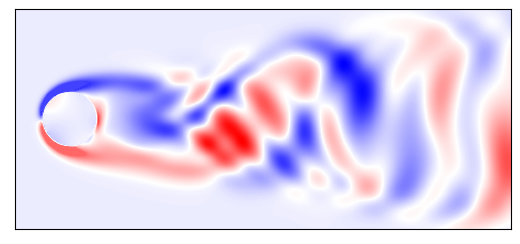}&
  \includegraphics[width=11em]{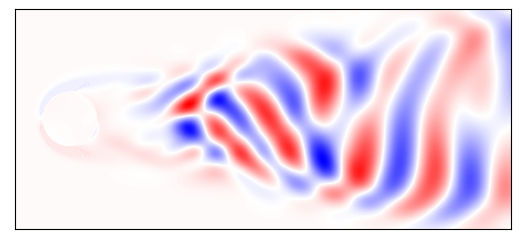}\\
  \multicolumn{4}{c}{(c) Our method with small data}\\
  \includegraphics[width=11em]{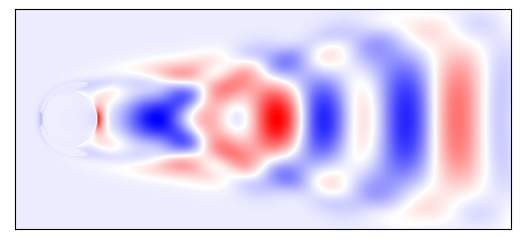}&
  \includegraphics[width=11em]{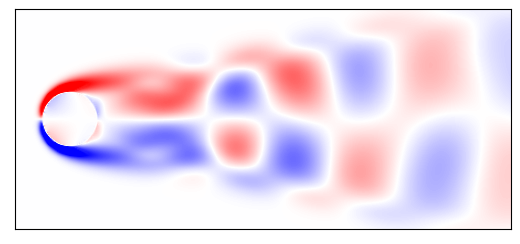}&
  \includegraphics[width=11em]{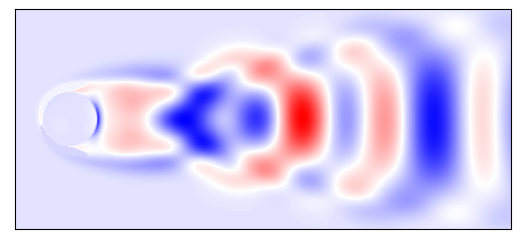}&
  \includegraphics[width=11em]{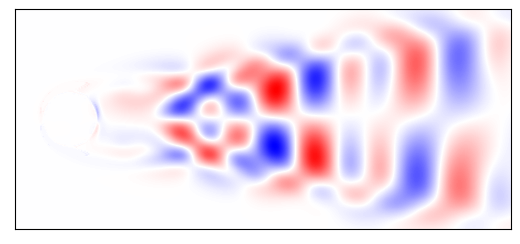}\\
  \end{tabular}}
  \caption{Dynamic modes by (a) DMD with large data, (b) DMD with small data, and (c) the proposed method with small data.}
  \label{fig:mode}
\end{figure*}

\subsection{Data}

To evaluate the proposed method,
we used the following six time-series data sets:
simple gravity pendulum (Pendulum),
Pendulum with irregularly spaced observation times (PendulumI),
Van der Pol oscillator (VanDerPol), fluid flow (Fluid),
sea surface temperature (SST), and bike share data (Bike).
These datasets have been used for evaluating nonlinear time-series models~\cite{azencot2020forecasting,lusch2018deep,greydanus2019hamiltonian,mauroy2019koopman,kutz2016dynamic,yang2019short}.
For all datasets,
a single short sequence was used for training to evaluate the proposed method
when a small number of observations are given.
We used the first 20\% of a sequence as the training data,
the following 10\% as the validation data,
and the remaining as the test data.
For each dataset,
the performance was evaluated by averaging the results of
30 experiments with different random seeds for generating the data.
The details of the datasets were described in the supplemental material.

\subsection{Compared methods}

We compared the proposed method
with the following methods:
NDMD, CKA, DMD, DMDF, FFNN, FFNNF, LSTM, LSTMF, LEM, NODE, HNN, and GHNN.
NDMD is a neural dynamic mode decomposition~\cite{takeishi2017learning}.
It corresponds to the proposed method without
eigen decomposed structured Koopman matrix,
where the information on decay rates or frequencies
cannot be incorporated into the model.
CKA is consistent Koopman autoencoders~\cite{azencot2020forecasting}.
It is an extension of NDMD,
where backward dynamics is modeled as well as forward dynamics,
and a regularizer that promotes consistent dynamics is introduced.
DMD is dynamic mode decomposition~\cite{williams2015data,rowley2009spectral,schmid2010dynamic,mezic2005spectral}.
It corresponds to NDMD without encoders and decoders,
where the linear dynamics in the measurement space is assumed.
DMDF is the dynamic mode decomposition with decay rate and frequency information,
which corresponds to the proposed method (OursF) without encoders and decoders.
FFNN is a feed-forward neural network, LSTM is
the long shot-time memory~\cite{hochreiter1997long},
and LEM is the long expressive memory~\cite{rusch2021long}.
They take a measurement vector and a prediction time period
as input, and outputs a predicted measurement vector
after the prediction time period from the given measurement vector,
by which they can be trained with irregularly sampled time-series.
FFNNF and LSTMF are FFNN and LSTM with frequency information,
where they use regularizers that make the prediction takes the same values
periodically with the given frequency. 
NODE is neural ordinary differential equations~\cite{chen2018neural},
where an ODE is modeled by a neural network,
and the output of the model is computed using an ODE solver.
HNN is Hamiltonian neural networks~\cite{greydanus2019hamiltonian},
which models the Hamiltonian by a neural network.
Given a set of coordinates,
which consist of the positions of objects and their momentum,
it can predict time derivatives of the coordinates.
For training HNN, we approximated the time derivatives by the finite difference. 
GHNN is weak form generalized Hamiltonian learning~\cite{course2021weak},
which models a generalized Hamiltonian decomposition~\cite{sarasola2004energy}
by a neural network.
The weak form learning allows one to drop the requirement of
approximating time derivatives
without having to backpropagate
through an ODE solver or solving an adjoint ODE,
where quadrature techniques are used
assuming the time measurements are sufficiently close together.
CKA, DMDF, HNN, and GHNN use the information on conservation laws.
DMDF, FFNNF, and LSTMF use the information on frequencies.

\subsection{Settings}

In the proposed method,
we used a three-layered feed-forward neural network
with four hidden units and two output units.
The dimensionality of
the Koopman space was $K=2$.
We fixed the real parts of the eigenvalues equal to zero, $r_{k}=0$.
With LSTM, LSTMF and LEM, a neural network was used to output the prediction
taking the hidden units of LSTM and LEM as input.
Their number of hidden units was 32.
With NDMD, CKA, FFNN, FFNNF, LSTM, LSTMF, LEM, NODE, and GHNN,
a four-layered feed-forward neural network was used.
The numbers of hidden units were selected
from $\{(4,2),(8,4),(16,8),(32,16)\}$ using the validation data.
With the proposed method, NDMD, DMD, DMDF, and CKA,
the minimum and maximum numbers of prediction steps
in the training loss were $\nu_{\mathrm{start}}=-10$ and $\nu_{\mathrm{end}}=10$.
With FFNN, FFNNF, LSTM, LSTMF and LEM,
they were $\nu_{\mathrm{start}}=1$ and $\nu_{\mathrm{end}}=10$.
The activation function in the neural networks
was the hyperbolic tangent.
Optimization was performed using Adam~\cite{kingma2014adam}
with learning rate $10^{-2}$.
The maximum number of training epochs was 5,000,
and the validation data were used for early stopping.
We implemented all methods with PyTorch~\cite{paszke2017automatic}.

\subsection{Results}

Table~\ref{tab:error} shows the test mean squared error,
where Ours is the proposed method with the known decay rate,
and OursF is the proposed method with the known decay rate and known frequency.
The proposed method achieved the smallest errors on all the datasets.
When the frequency information was provided,
the proposed method improved the performance.
Figure~\ref{fig:vanderpol} shows the predicted phase space
and time-series on VanDerPol data by OursF, NDMD, NODE, and HNN.
OursF successfully modeled preserving measurements
as shown by the predicted trajectories in the phase space
in Figure~\ref{fig:vanderpol}(a).
Figure~\ref{fig:vanderpol_g} shows the encoded time-series in the Koopman space
on VanDerPol data by OursF.
Although the dynamics in the measurement space was nonlinear as shown in Figure~\ref{fig:vanderpol}(a),
that in the Koopman space was linear, which can be represented by a sine wave.

Since NDMD does not have constraints on the decay rates,
the trained model exhibited the diverse dynamics as in shown Figure~\ref{fig:vanderpol}(b).
CKA improved the performance compared with NDMD except for PendulumI data
due to the regularizer for consistent dynamics.
However, since the regularizer is not exact constraints,
the measurement was not perfectly conserved,
and the long-term prediction was worse than the proposed method.
Since DMD assumes linear dynamics in the measurement space,
it failed to model the nonlinear dynamics.
The forecasting performance by FFNN, LSTM, and LEM was worse than the proposed method
since they cannot use inductive bias.
DMDF, FFNNF, and LSTMF improved the performance by using the frequency information
although they underperformed the proposed method.
Since NODE did not use the decay rate information,
it mistakenly trained decaying dynamics
although the short-term prediction error was small as shown in (c).
With HNN and GHNN, the trained dynamics followed conservation laws.
However, their long-term prediction errors were higher than the proposed method
as shown in (d) and Table~\ref{tab:error}.
It is because both of the methods require that
the time measurements are sufficiently close together,
where HNN uses the time derivatives approximated by
the finite difference, and GHNN uses quadrature techniques.
In Table~\ref{tab:error},
the performance by Ours, NDMD, and NODE on PendulumI data
was better than that on Pendulum data.
It is because Ours, NDMD, and NODE predict measurement vectors
by calculating the integration over time,
and they are trained with various time intervals on PendulumI data.
In contrast,
since FFNN, LSTM, LEM, HNN, and GHNN do not explicitly
calculate the integration, their performance on PendulumI data
was worse than that on Pendulum data.

Table~\ref{tab:ablation} shows the test mean squared error by the proposed method
without a backcast loss for training (w/o backcast) and without structured eigenvectors (w/o struct eigen)
on VanDerPol data.
With the w/o backcast, the start number of prediction steps was set to zero, $\nu_{\mathrm{start}}=0$,
for the training objective function in Eq.~(\ref{eq:E}).
With the w/o struct eigen,
each element of eigenvectors $\vec{V}$ of the Koopman matrix was considered as a parameter to be trained,
where structured eigenvectors in Eq.~(\ref{eq:v}) were not used.
The better performance with backcast and with structured eigenvectors indicates their effectiveness.

Table~\ref{tab:range} shows the test mean squared error by the proposed method in the cases that ranges of frequencies were given, where we used the sigmoid function as in Eq.~(\ref{eq:range}) for specifying the range of frequencies. 
As the range width was shortened, the proposed method improved the performance.
We evaluated the estimated frequencies by the proposed method with the known decay rate, which is shown in Table~\ref{tab:freq}. Using the decay rate constraints, the proposed method achieved better frequency estimation performance than the other Koopman-based methods. In addition, we estimated an unknown frequency given a known frequency using a system with two frequencies, where the data were generated by adding a sine wave with frequency two to VanDerPol data. Table~\ref{tab:freq2} shows that 
the proposed method that used the sine wave frequency information (OursF) improved the estimation performance of the unknown frequency.

Figure~\ref{fig:mode} shows
dynamic modes by DMD with large data, DMD with small data, and the proposed method with small data.
Dynamic modes represent synchronization patterns,
and they are calculated by decoded left eigenvectors $\phi(\vec{V}^{-1})$~\cite{tu2013dynamic}.
The large data used the first 80\% of the vorticity time-series in $449\times 199$ fields
for training.
The small data used the first 20\% of the time-series.
In the proposed method, we used a linear encoder and decoder based on singular value decomposition.
In all methods, the dimensionality of the Koopman space was eight.
The dynamic modes with large data are ideal.
The proposed method extracted clearer patterns than
DMD with small data using the inductive bias on the decay rate
even though they used the same data.
More experimental results are in the supplemental material.

\section{Conclusion}
\label{sec:conclusion}

We proposed a method for modeling nonlinear dynamical systems in continuous time.
The proposed method embeds observations to a Koopman space with neural networks,
by which we can forecast and backcast effectively imposing information on decay rate and frequency
due to the linearity of the dynamics in the Koopman space.
Although we believe that our work is an important step for modeling nonlinear dynamics with inductive bias,
we must extend our approach in several directions.
First, we will extend our method such that it can handle dynamics with time-varying frequencies
by incorporating~\cite{lusch2018deep} in our framework.
Second, we want to incorporate non-periodic dynamics in our model as well as periodic dynamics.

\bibliography{arxiv_param_dmd}
\bibliographystyle{abbrv}

\end{document}